\documentclass{article}
\usepackage{spconf,amsmath,graphicx}
\usepackage{todonotes}
\usepackage{url}
\usepackage{listings,fancyvrb}
\usepackage[]{mcode}

\newenvironment{mitemize}{
  \begin{itemize}
    \setlength{\itemsep}{1pt}
    \setlength{\parskip}{0pt}
    \setlength{\parsep}{0pt}}{\end{itemize}
}

\title{Blob indentation identification via curvature measurement}
\name{Matthew J. Sottile}
\address{Sailfan Research, Inc.\\
         Portland, OR, USA\\\url{mjsottile@gmail.com}}
\begin{document}
%
\maketitle
\begin{abstract}
  This paper presents a novel method for identifying indentations on
  the boundary of solid 2D shape.  It uses the signed curvature at a
  set of points along the boundary to identify indentations and
  provides one parameter for tuning the selection mechanism for
  discriminating indentations from other boundary irregularities.  An
  efficient implementation is described based on the Fourier transform
  for calculating curvature from a sequence of points obtained from
  the boundary of a binary blob.
\end{abstract}
\section{Introduction}
\label{sec:intro}

Blob detection is a common unit in image processing pipelines in which
an image is reduced to a set of distinct solid connected components
that are then analyzed further.  Most image processing packages
provide a number of useful geometric and topological quantities that
can be computed for blobs, such as their centroid, area,
eccentricity, perimeter, and so on.  In some applications it is
important to look at properties of the boundary of the blobs.  In
particular, indentations and irregularities in the shape boundary that
may have specific meaning in a given domain (e.g., irregularly shaped
cells in microscopy, defective widgets in manufacturing, etc...).  How
does one go about characterizing these indentations and irregularities
to count them, locate them, and measure them?

In this paper a method is presented to analyze blob boundaries to
identify indentations based on deriving a parameterized curve
enclosing the blob and using measures of curvature on this curve to
determine where indentations are present.  A trigonometric polynomial
representation of the boundary curve is obtained via the Fourier
transform, and the necessary derivatives to compute the signed
curvature are obtained using convenient properties of the Fourier
transform that make such computations very simple to state and
implement.  This method requires one parameter, \emph{severity}, for
tuning what constitutes an indentation versus a simple irregularity in
the shape.  This parameter has a geometric interpretation related to
the radius of curvature along the boundary giving indentation
selection a natural length scale.  The method will be briefly compared
with other approaches to this problem using common algorithms such as
the convex hull of the shape to demonstrate where this method is
preferable.


\section{Approach}
\label{sec:format}

The key tool used in this method is the interpretation of the boundary
of the shape as a parametric curve from which one can compute the
approximate curvature at a sequence of points.  When the sign of
the curvature changes (i.e., when the normal to the curve switches
from pointing into the shape to outside, or vice versa), one can infer
that an inflection point has occurred corresponding to an indentation
beginning or ending.  The steps of the algorithm are:

\begin{mitemize}

\item Derivation of a trigonometric polynomial representation of the
  discrete blob boundary point sequence as a continuous curve
  parameterized by arc length from an arbitrary boundary starting
  position.

\item Optional low-pass filtering of the parameterized boundary curve.
  This step eliminates high-frequency oscillations in the curve due to
  the discretization of the continuous shape when pixelized.

\item Calculation of the first and second derivative along the curve
  by element-wise multiplication in the frequency domain.
  Reconstruction of the first and second derivative in the spatial
  domain necessary to calculate the curvature along the boundary is
  obtained via the inverse Fourier transform.

\item Identification of inflection points along the curve at points of
  curvature sign change.  The regions delimited by these inflection
  points are then filtered by radius of curvature to eliminate
  detection of non-indentation features.

\end{mitemize}

A prerequisite for this algorithm is an existing segmentation pipeline
that yields binary blobs from images, as well as a method for
extracting an ordered point sequence along the boundary of these
blobs.  Such methods exist in many off-the-shelf software packages
(such as the MATLAB image processing toolbox).

\subsection{Related approaches}

Shape analysis is a widespread activity in image processing.  The
majority of techniques found in the literature or published software
packages rely on working directly with the pixel boundary of a shape
or a polygon approximation for either the shape or its convex hull.
The use of the convex hull for providing an approximation for a shape
from which other properties can be easily computed is popular in large
part due to the algorithmic efficiency and widespread availability of
convex hull algorithms in popular packages using the QuickHull
method~\cite{Barber96thequickhull,preparata85geometry}.  As discussed
later in Section~\ref{sec:convexhull}, these methods make significant
assumptions about what constitutes a shape indentation limiting their
utility.

Approaches that do not make assumptions about how indentations relate
to the convex hull of a shape are more robust to diverse shapes.
Alpha shapes~\cite{edelsbrunner83shape,edelsbrunner10topology} and
Delaunay triangulations~\cite{dey07curve} can be used to calculate a
boundary of an image from a point set (e.g., a set of pixels
identified along the boundary of a binary blob).  These methods
unfortunately still require heuristic tests to determine when an
indentation occurs given that the decision as to whether or not an
indentation is present requires more information than can be derived
from a single boundary line segment or vertex.

The calculation of the curvature using a Fourier series to represent
the boundary curve naturally takes into account all points along the
boundary.  As such, the calculation of curvature at any point along the
curve is intrinsically informed by the properties of the entire curve
removing the need for any local heuristics near the point where
curvature is calculated.  This simplifies the algorithm by removing
heuristics that would induce additional parameters and potential
points of brittleness.

\section{Curvature calculation}
\label{sec:pagestyle}

Given a blob with a corresponding boundary pixel sequence $B=((x_1,
y_1), ..., (x_n, y_n))$, the goal is to calculate the curvature
$\kappa$ at each point along $B$.  The signed curvature can be defined
at the $i$th point along the curve in terms of the local first and second
derivatives at that point:

\begin{equation}
  \label{eq:kappa}
  \kappa_i = \frac{x_i'y_i''-y_i'x_i''}{\left( {x_i'}^2 + {y_i'}^2 \right)^\frac{3}{2}}
\end{equation}

The derivatives along $B$ must be calculated from the point sequence
that was derived from the blob boundary.  A simple and fast method for
achieving this is via the Fourier transform~\cite{tolstov67fourier}.
The Fourier transform has a very convenient property that allows for
simple calculations of derivatives of any function represented as a
Fourier series.  Given some function $x(t)$, the following property
holds for its derivatives:

\begin{equation}
  \label{eq:differential-operator}
  \mathcal{F} \left[ \frac{d^n}{dt^n} x(t) \right] = (i \omega)^n X(i \omega)
\end{equation}

Where $X(i \omega) = \mathcal{F} [x(t)] $.  This allows the function
to be differentiated in the frequency domain via simple
multiplication.  In the case of blob boundaries, the function $x(t)$
can be replaced by the sampled boundary points $x[t]$.  The $n$th
derivative in the spatial domain necessary for the calculation of
$\kappa$ is recovered via the inverse Fourier transform:

\begin{equation}
  \label{eq:diff-inverse-fft}
  x^{(n)}(t) = \mathcal{F}^{-1}\left[ (i \omega)^n X(i \omega) \right]
\end{equation}

This approach is attractive for a number of reasons.  First, the
Fourier representation does not use a spatially local approximation
for the curve in calculating the derivatives.  This eliminates the
need to select parameters such as a stencil size when using finite
difference approximations for calculating derivatives directly from
the boundary point sequence.  Second, once the boundary curve has been
moved to the frequency domain additional processing is possible to
address issues such as high-frequency oscillations along a boundary
due to pixel-level effects.  Application of low-pass filters is very
simple in the frequency domain and composes cleanly with the
calculation of the derivative.  In fact, any additional processing of
the shape boundary that can be represented via convolutions can be
included in the calculations applied to the boundary while in the
frequency domain.

\begin{figure}[t]
  \begin{minipage}[b]{1.0\linewidth}
    \centering
    \centerline{\includegraphics[width=7cm]{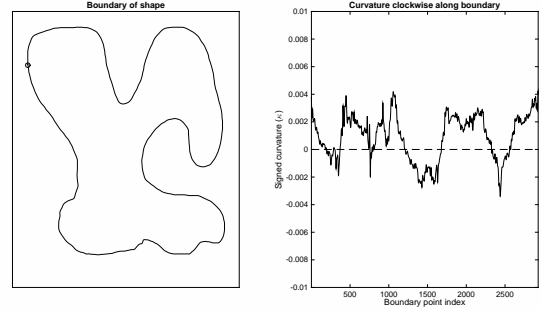}}
  \end{minipage}
  \caption{Illustration of a parameterized curve boundary and the corresponding curvature measure along the boundary (clockwise, starting at indicated point on boundary.)}
  \label{fig:exampleblob}
\end{figure}

\section{Indentation identification}
\label{sec:dentcounting}

Identifying indentations on a blob boundary using curvature requires
searching for inflection points where the normal to the curve changes
sign.  This occurs at points where the curvature reaches zero.
Consider the example blob in Figure~\ref{fig:exampleblob}.  The
curvature along the curve is shown in the right-hand plot, showing
clearly the points along the curve where the curvature passes through zero.

One parameter is used to tune indentation selection, \emph{severity}
($\sigma$).  This parameter has a close relationship with the radius
of curvature that is used to define what constitutes an indentation.
The magnitude of curvature, $|\kappa|$, at a point along the boundary
can be related to the radius $\rho$ of the circular arc that fits the
curve at that point~\cite{henderson98differential}:

\begin{equation}
\label{eqn:curvprop}
\frac{1}{|\kappa|} \propto \rho
\end{equation}

\begin{figure}[t]
  \begin{minipage}[b]{1.0\linewidth}
    \centering
    \centerline{\includegraphics[width=5.5cm]{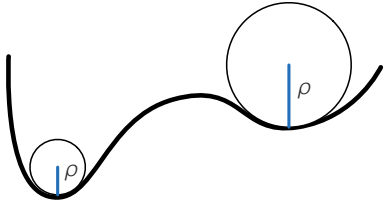}}
  \end{minipage}
  \caption{An illustration for the radius of curvature for different sized indentations.}
  \label{fig:radcurv}
\end{figure}

In regions of very low magnitude curvature, the corresponding circular
arc would have a very large radius.  Similarly, when the magnitude of
curvature is high, the best-fit circular arc would have a low radius.
Indentations can be thought of as places on the boundary where the
sign of curvature is negative and the magnitude of curvature is high,
and the use of the radius of curvature can provide a cutoff length
scale for filtering indentations from non-indentation irregularities
in which the sign of curvature is also negative.

Given a maximum radius of curvature for what would be considered an
indentation, $\rho_\sigma$, then $\rho < \rho_\sigma$ can be
interpreted with respect to curvature by Eq.~\ref{eqn:curvprop} as
$|\kappa|^{-1} < \sigma^{-1}$.  Thus indentations will be considered
only when $|\kappa| > \sigma$.

\section{Algorithm}
\label{sec:algorithm}

The algorithm for the curvature calculation can be specified in MATLAB
as follows:

\begin{lstlisting}
function k = curvature(xs, ys)
  fxs = fft(xs);  fys = fft(ys);

  nx = length(xs);
  hx = ceil(nx/2)-1;
  ftdiff = (2i*pi/nx)*(0:hx);
  ftdiff(nx:-1:nx-hx+1) = -ftdiff(2:hx+1);
  ftddiff = (-(2i*pi/nx)^2)*(0:hx);
  ftddiff(nx:-1:nx-hx+1) = ftddiff(2:hx+1);

  dx = real(ifft(fxs.*ftdiff'));
  dy = real(ifft(fys.*ftdiff'));
  ddx = real(ifft(fxs.*ftddiff'));
  ddy = real(ifft(fys.*ftddiff'));

  k = (ddy.*dx-ddx.*dy) ./ ((dx.^2+dy.^2).^(3/2));
end
\end{lstlisting}

First, the x and y coordinate arrays of the points along the curve are
mapped to the frequency domain with the FFT.  The differential
operator is calculated based on the length of the point sequences
based on Eq.~\ref{eq:differential-operator}.  This operator is then
applied by pointwise multiplication against the frequency
representation of the points and the first and second spatial
derivatives are then obtained by taking the real component of the
inverse Fourier transform of the products
(Eq.~\ref{eq:diff-inverse-fft}).  The calculation of $\kappa$ then
follows exactly as defined in Eq.~\ref{eq:kappa}.

Identification and counting of indentations then requires processing
of the curvature sequence, as performed in the following pseudocode.

\begin{lstlisting}
function n = countindentations(k, sigma)
  ksigns = sign(k);
  ksignchanges = diff(ksigns);
  idxchanges = [1; find(ksignchanges); ...
                length(ksignchanges)];

  n = 0;
  for i = 1:length(idxchanges)-1
    ks = k(idxchanges(i):idxchanges(i+1));
    if mean(ks) < 0 && max(abs(ks)) > sigma
      n = n + 1;
    end
  end

  return n
end
\end{lstlisting}

This algorithm takes the curvature sequence \texttt{k} and identifies where
sign changes occur.  This is used to partition the curvature sequence into a
set of regions with the same sign curvature.  Each is then examined to
identify those that represent negative curvature regions in which the
maximum curvature exceeds the value of the severity parameter \texttt{sigma}.

\subsection{Demonstration}

\begin{figure}[t]
  \begin{minipage}[b]{1.0\linewidth}
    \centering
    \centerline{\includegraphics[width=7cm]{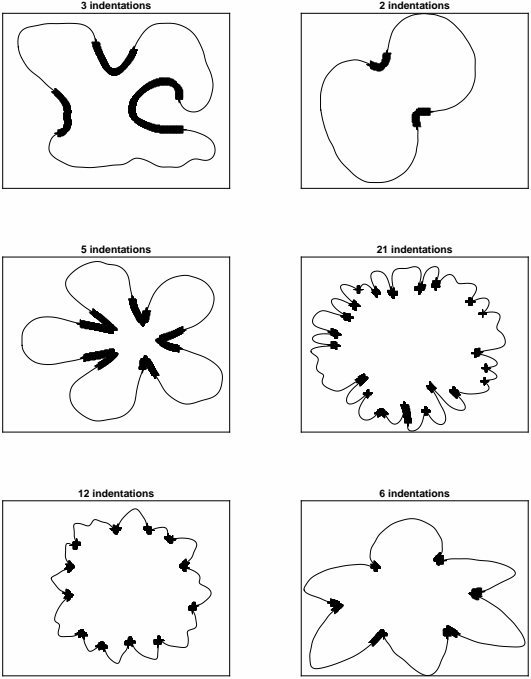}}
  \end{minipage}
  \caption{Demonstration of the algorithm for six test images: two synthetic and four obtained from segmented photographs.}
  \label{fig:demo}
\end{figure}

The application of this algorithm is shown in Figure~\ref{fig:demo} for
six different shapes: two synthetic blobs drawn by hand, and four blobs derived
from photographs of different types of flowers.  In each of the sub-images
the regions of negative curvature are indicated by the heavy boundary curve,
and the title indicates the number of indentations that were detected.
In some cases the effect of the $\sigma$ parameter is apparent where
irregularities along the surface were ruled out as indentations since their
maximum curvature did not exceed the threshold corresponding to the
acceptable range of radii of curvature for indentations.

\subsection{Comparison with convex hull approaches}
\label{sec:convexhull}

\begin{figure}[t]
  \begin{minipage}[b]{1.0\linewidth}
    \centering
    \centerline{\includegraphics[width=6.5cm]{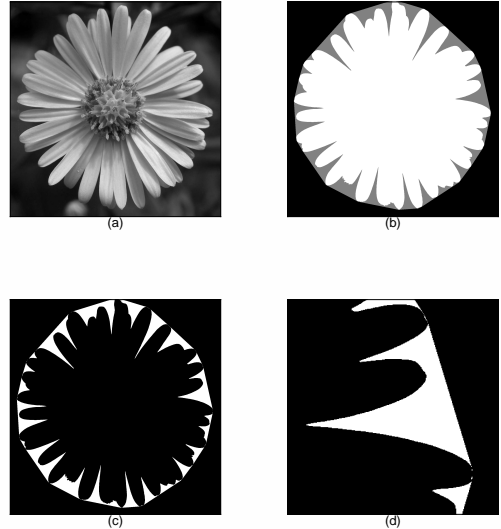}}
  \end{minipage}
  \caption{An example showing indentations missed by examining the difference between the shape and its convex hull.}
  \label{fig:convexhull}
\end{figure}

A simple approach to solving this same problem using readily available
algorithms is to examine the convex hull of the shape and count the
number of regions where a gap appears between the convex hull and the
curve itself.  An example is shown in Figure~\ref{fig:convexhull}(a),
with the blob and its convex hull shown in
Figure~\ref{fig:convexhull}(b).  Gaps between the shape and the convex
hull can be obtained by subtracting the blob from its convex hull as
shown in Figure~\ref{fig:convexhull}(c).  The problem with this
approach is that it makes a simplistic assumption about what
constitutes an indentation: specifically, that a shape without
indentations will strictly match its own convex hull.  This is
unrealistic in many applications in which non-indentation boundary
irregularities are expected.  A consequence of this is that some
indentations may be missed: a detailed region along the shape boundary
is shown in Figure~\ref{fig:convexhull}(d).  Using the calculated
curvature along the shape boundary is insensitive to this kind of shape
since the curvature analysis makes no assumptions about the shape
relative to a canonical analogue such as the convex hull.

The attraction of convex hull methods is the ready availability of
efficient algorithm implementations.  Fortunately, the method
presented in this paper also leverages a fundamental algorithm with
widespread deployment, the Fast Fourier Transform.  As such, the methods
described in this paper are competitive in both accessibility as well as
algorithmic efficiency since the FFT has time complexity $O(n \log n)$ for
$n$ points, comparable to the average time complexity of the QuickHull
method ($O(n \log n$) average, $O(n^2)$ worst-case).

\section{Conclusions}

In this paper a method has been demonstrated to use the computed
curvature along the boundary of a 2D shape to identify indentations.
This method is based on calculation of curvature via the Fourier
transform and selection of indentations using a single scalar
parameter corresponding to the maximum radius of curvature allowed to
be considered an indentation.  Unlike methods based on the convex hull
or piecewise linear polygon approximations, the selection of
indentations requires no heuristic tests and cleanly integrates
frequency-domain filtering of the shape boundary to eliminate
pixel-level discretization effects along the boundary of the shape.

An implementation of this algorithm is available as an open source MATLAB
package\footnote{\url{http://github.com/mjsottile/blobdents/}}.

\bibliographystyle{IEEEbib}
\bibliography{refs}

\end{document}